\def\year{2020}\relax
\documentclass[letterpaper]{article} % DO NOT CHANGE THIS
\usepackage{aaai20}  % DO NOT CHANGE THIS
\usepackage{times}  % DO NOT CHANGE THIS
\usepackage{helvet} % DO NOT CHANGE THIS
\usepackage{courier}  % DO NOT CHANGE THIS
\usepackage[hyphens]{url}  % DO NOT CHANGE THIS
\usepackage{graphicx} % DO NOT CHANGE THIS
\usepackage{graphicx}  % DO NOT CHANGE THIS
\usepackage{bm}
\usepackage{amsmath}
\usepackage{amsfonts}
\usepackage{comment}
\title{New Perspective of Interpretability of Deep Neural Networks}
\author{Masanari Kimura\textsuperscript{\rm 1}, Masayuki Tanaka\textsuperscript{\rm 1,2} \\% All authors must be % in the same font size and format. Use \Large and \textbf to achieve this result when breaking a line
 \textsuperscript{\rm 1}National Institute of Advanced Industrial Science and Technology\\
 \textsuperscript{\rm 2}Tokyo Institute of Technology\\
mkimura@aist.go.jp\\
mtanaka@sc.e.titech.ac.jp % email address must be in roman text type, not monospace or sans serif
}
 
\begin{document}

\maketitle

\begin{abstract}
Deep neural networks (DNNs) are known as black-box models. 
In other words, it is difficult to interpret the internal state of the model.
Improving the interpretability of DNNs is one of the hot research topics.
However, at present, the definition of interpretability for DNNs is vague, and the question of what is a highly explanatory model is still controversial.
To address this issue, we provide the definition of the human predictability of the model, as a part of the interpretability of the DNNs.
The human predictability proposed in this paper is defined by easiness to predict the change of the inference when perturbating the model of the DNNs.
In addition, we introduce one example of high human-predictable DNNs.
We discuss that our definition will help to the research of the interpretability of the DNNs considering various types of applications.
\end{abstract}

\section{Introduction}
% About DNNs, hyper-parameter tuning, etc. Why we have to interpret DNNs internal state.
In recent years, Deep Neural Networks (DNNs) have achieved great success in a number of tasks~\cite{deng2009imagenet,liu2017survey}.
Nevertheless, the internal state of DNNs are known to be difficult for humans to understand because of their complexity.
Due to these black-box properties of the DNNs, numerous iterations of trial-and-error are required to develop the DNN-based applications.
One approach to solve this problem is AutoML~\cite{elsken2018neural,zoph2016neural}.
AutoML is a field of machine learning concerned with automating repetitive tasks such as model selection or hyperparameter optimization.
However, AutoML is still a very challenging problem, especially when given complex data.

% The necessity of human insight for tuning DNNs.
In practice, human insight is essential to improve the performance of DNN-based applications. 
This kind of development involving human is called Human-In-The-Loop (HITL) machine learning~\cite{holzinger2016interactive,fails2003interactive}.%, which schematic is shown in Fig.~\ref{fig:human_in_the_loop}.
In the human-in-the-loop approach, people are involved in a virtuous circle where they train, tune, and test a particular algorithm. The goal of HITL machine learning is to train more desired models by human intervention compared to simple end-to-end training.
Interpretability and/or explainability of the DNN model is one of the critical aspects of the HITL machine learning.

So-called explainable AI (XAI) has become one of the hot topics in the machine learning field~\cite{gunning2017explainable,samek2017explainable}.
However, a big issue is that the definition of the interpretability and/or explainability of the DNN model is ambiguous.

\begin{figure}
  \centering
  \includegraphics[scale=0.37]{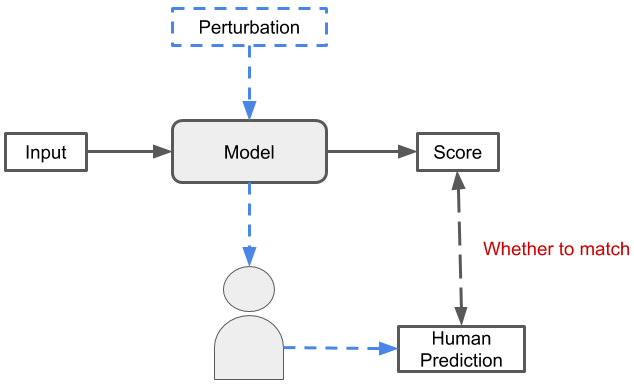}
  \caption{Overview of "human predictability of the model."
  We define human predictability as easiness for the human to predict the change of the inference when perturbing the model.
  \label{fig:human_predictable}}
\end{figure}

% proposed definition
To tackle this problem, we define "human predictability of the model" as "easiness for the human to predict the change of the inference when perturbing the model."
Figure~\ref{fig:human_predictable} shows the overview of our definition.
This means that even if the function is complex, the result of the operation on it must match human intuition.
In this paper, we show that some functions other than DNNs fit our definition.
Furthermore, we propose an example of a neural network architecture and operation pair that satisfies our definition.

\begin{comment}
\begin{figure*}
  \centering
  \includegraphics[width=16cm]{human_in_the_loop.png}
  \caption{Overview of Human-In-The-Loop structure.  In a human-in-the-loop approach, people are involved in a virtuous circle where they train, tune, and test a particular algorithm. The goal of this structure is to train more desired models by human intervention compared to end-to-end learning.\label{fig:human_in_the_loop}}
\end{figure*}
\end{comment}

\section{Related Works}

\subsection{Humain-In-The-Loop Machine Learning}
Human-in-the-loop (HITL) is a research topic of machine learning scheme that leverages both human and machine to create intelligent models.
One of the essential paradigms of HITL machine learning is active learning~\cite{settles2009active}.
The active learning is a framework that allows humans to select and input training data that maximizes the performance of the next training-step model based on the current model's output and dataset information.
There are several different problem scenarios in active learning.

For example, if we focus on the training data that humans can observe, we can classify them into stream-based and pool-based active learning.
In the stream-based active learning~\cite{dasgupta2008general}, humans decide whether to use the training data sampled each time.
The stream-based active learning scenario has been studied in several real-world tasks, such as part-of-speech tagging~\cite{dagan1995committee} or sensor scheduling~\cite{krishnamurthy2002algorithms}.

In the pool-based active learning~\cite{lewis1994heterogeneous}, a set of candidate training data is given in advance, and humans select candidate subsets based on the current model output.
The pool-based active learning is the major problem setting and has been studied for many tasks, such as image classification~\cite{tong2001support2} and speech recognition~\cite{tur2005combining}.

The main difference between the stream-based and the pool-based active learning is that the stream-based active learning obtains the data sequentially and makes query decisions individually, whereas the pool-based active learning evaluates the entire dataset before selecting the best query.
In this paper, humans intervene in the context of perturbing to DNNs during inference.

\begin{figure*}
  \centering
  \includegraphics[width=16cm]{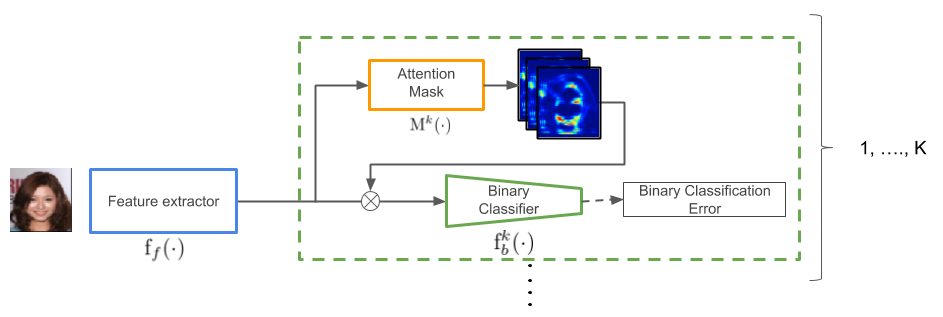}
  \caption{Overview of proposed network architecture.
  Here, there are $K$ binary classification components, where $K$ is the number of attributes.\label{fig:network_overview}}
\end{figure*}

\subsection{Interpretability of DNNs}
Currently, several analyses of the machine learning model's interpretation have been proposed.
We overview some of them.

\subsubsection{Sensitivity Analysis}
Sensitivity analysis~\cite{baehrens2010explain,simonyan2013deep} is a method to explain a prediction based on the model's locally evaluated gradient.
Sensitivity analysis quantifies the importance of each input variable $i$ as
\begin{equation}
    \label{eq:sensitivity_analysis_1}
    \mathcal{R}_i^s = \|\frac{\partial}{\partial x_i} f(x) \|.
\end{equation}
This measure assumes that the more relevant input features are those to which the output is more sensitive.

\subsubsection{Layer-Wise Relevance Propagation}
Layer-wise relevance propagation (LRP)~\cite{bach2015pixel} explains the classifier’s decisions by decomposition.
LRP redistributes the prediction backward via local redistribution rules until it assigns a relevance score to each input variable.
Mathematically, the redistribution can be summarized as
\begin{equation}
    \label{eq:lrp_1}
    \sum_i \mathcal{R}_i^l = \dots = \sum_j \mathcal{R}_j^l = \dots = f(x)
\end{equation}
The relevance scores $\mathcal{R}_i^l$ define how much the respect variable has contributed to the prediction. 

\subsubsection{Visual Explanation with Attention Mechanism}

There are many studies on visual explanation using attention mechanism~\cite{zhou2016learning,chattopadhay2018grad,fukui2019attention,kimura2019interpretation}.

CAM~\cite{zhou2016learning} allows us to visualize attention maps for each class using the response of a convolution layer and the weight at the last fully-connected layer.
CAM can obtain a good result for weakly-supervised object localization problem but not as well in image classification due to replacing fully-connected layers with convolution layers and passing through global average pooling layer.

Grad-CAM~\cite{simonyan2013deep} and Grad-CAM++~\cite{chattopadhay2018grad} are the gradient-based
visual explanation methods.
They can visualize an attention map using positive gradients at a specific class in backpropagation.

\subsubsection{Feature Importance}
Feature importance approach assumes the importance of a feature by calculating the increase in the model's prediction error after permuting the feature~\cite{fisher2018model}. 
Permutation feature importance does not require retraining the model or any additional module such as attention mechanism or global average pooling.

\subsubsection{Surrogate of Black-Box Model}
Surrogate models are trained to approximate the predictions of the underlying black-box model.

LIME~\cite{ribeiro2016should} focuses on training local surrogate models of the black-box models to explain individual predictions.
LIME examine what happens to the predictions when inputting the data into the target model. LIME generates a new dataset consisting of permuted samples and the corresponding predictions of the black-box model. 
On this new dataset, LIME then trains an interpretable model. The new model is weighted by the proximity of the sampled instances to the instance of interest. 

\section{Human Predictability of the Model}
The interpretability of the DNNs includes several concepts. The definition of the interpretability of the DDNs is still unclear. In this paper, we introduce the concept of the human predictability of the model, which is closely related to the interpretability. Here, we first provide the definition of the human predictability of the model. Then, we explain it with simple regression models.

\subsection{Definition}
The partial dependence plot and the feature importance are widely used to analyze the relationship between the inference and the input or the feature. 
Those analyses basically observe the inference while perturbating the input or the feature. 
The reason why those kinds of numerical analyses are required is that the human cannot predict the tendency of the change of the inference when perturbating the input or the feature. 
In this sense, if the human can easily predict the tendency of the change of the inference when perturbating the input or the feature of the model, we can say that the model has high interpretability. 
Based on those observations, in this paper, we define the human predictability of the model by "easiness to predict the change of the inference when perturbating the model." In the next subsection, we explain our definition of the human predictability of the model with simple regression models.

\subsection{Example of Simple Regression Models}
Here, we consider simple regression problem for given data $\{ (x_i,y_i) \}$, where $-1 \le x_i \le 1$, with three different regression models: (1) naive polynomials, (2) Legendre polynomials, and (3) Fourier series. 
The naive polynomial model can be expressed as
\begin{eqnarray}
 f_1(x) &=& \sum_{n=0}^{2N} a_n x^n \,,
 \label{eq:NaivePoly}
\end{eqnarray}
where $a_n$ is coefficient.
In some senses, the naive polynomial model is straightforward and easy to understand.
However, as we discuss later, from the point of view of our definition, the human predictability of the naive polynomial model is not so good, because each term of the naive polynomial is not orthogonal each other.

Legendre polynomials is well-known orthogonal basis, which can be expressed as
\begin{eqnarray}
 f_2(x) &=& \sum_{n=0}^{2N}
 a_n P_n(x) \,,
 \label{eq:NaivePoly2}
\end{eqnarray}
where
\begin{eqnarray}
 P_n(x) &=& 2^n \sum_{k=0}^{n} x^k 
 \left(
  \begin{array}{c}
    n \\
    k 
  \end{array}
 \right)
 \left(
  \begin{array}{c}
    \frac{n+k-1}{2} \\
    n 
  \end{array}
 \right)
 \,, \\
 \left(
  \begin{array}{c}
    n \\
    k 
  \end{array}
 \right)
 &=&
 \frac{n!}{k!(n-k)!}
 \,,
\end{eqnarray}
where $n!$ is the factorial of $n$.
The Legendre polynomial model seems to be complicated compared to the naive polynomial model in Eq.~\ref{eq:NaivePoly}.
However, it can be considered that the Legendre polynomial model has higher human predictability compared to the naive polynomial model because each term of the Legendre polynomial model is orthogonal each other.
That property means that the modification of the coefficient does not affect other terms.

A Fourier series is also known as orthogonal basis, which is expressed as
\begin{eqnarray}
 f_3(x) &=&
 \frac{a_0}{2}
 \sum_{n=1}^{N}
 a_n \cos \pi n x + b_n \sin \pi n x
 \,,
\end{eqnarray}
where $a_n$ and $b_n$ are coefficients.

Now, we compare the human predictability of those three models.
Let us consider perturbation for the coefficient of each model. 
For the Fourier series, the frequency component associated with the perturbed coefficient is only affected. 
It is straightforward to predict for human. 
The Legendre polynomial has similar properties to the Fourier series because the Legendre polynomial is the orthogonal basis.
The difference between the Legendre polynomial and the Fourier series is the meaning of the basis.
The basis of the Legendre polynomial is difficult to imagine for the human, while the base of the Fourier series is easy to imagine for the human because the basis of the Fourier series corresponds the wave with a specific frequency. 
From those properties, we can say the Fourier series has higher human predictability than the Legendre polynomial.
The naive polynomial is the lowest human predictability among the three models because the basis is not orthogonal.
It means that the perturbation of the coefficient affects other components.
It is difficult to predict the change for human.

\begin{figure}[tb]
  \centering
  \includegraphics[scale=0.28]{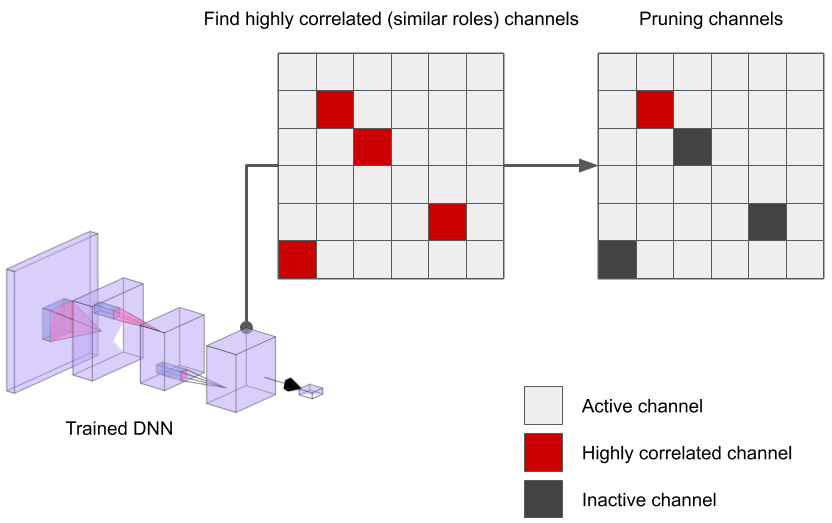}
  \caption{Semantic Channel Pruning operation.
  We sequentially scan the highly correlated channel pairs and prune the channel with the lower $L1$-norm of the weights.\label{fig:feature_semantic_pruning}}
\end{figure}

\section{Human Predictable DNNs}
\begin{comment}
In this section, we introduce one example of human-interpretable DNNs and the corresponding operations of making perturbation "Semantic Channel Pruning" and "Intentional Attention Mask Transformation."
\end{comment}
In this section, we introduce one example of human interpretable DNNs  which have high human interpretability for the perturbation of the operations of ``Semantic Channel Pruning'' and ``Intentional Attention Mask Transformation.''
Figure~\ref{fig:network_overview} shows the outline of our proposed network architecture.

\begin{comment}
Our network architecture takes multi-attributed images as input and reveals the correlation in terms of the role among the weights in the feature space.
\end{comment}
Our network basically recognizes multi-attribute for the input image.
It is easy to obtain the correlation between features.
Our proposed network architecture consists of the following two modules.
\begin{itemize}
    \item{Feature Extractor}
    \item{Binary Classifiers with Channel Attention}
\end{itemize}
We aim to obtain highly interpretable neural networks that satisfy our definition using an attention mechanism to acquire features that are important for classifiers.

\subsection{Feature Extractor}
Feature extractor $f_f$ extracts generic feature used by all the following networks. This component performs feature extraction. For this component, we use the dilation network \cite{fisher2016multi}.
Table~\ref{table:feature_extractor} shows the detailed architecture of the feature extractor that we used in the experiment.

\begin{figure*}[t]
  \centering
  \includegraphics[scale=0.56]{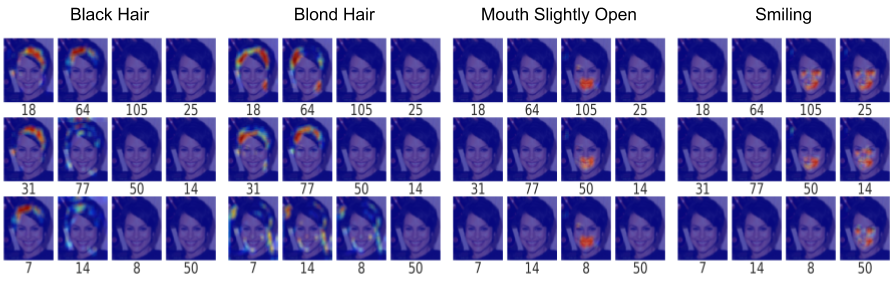}
  \caption{Visualizing attention masks on multiple facial attributes recognition. One element is one channel of the attention mask. The number under each mask means feature IDs. It can be seen that even in different channels, there are cases where attention focus to the same location in the image.}
  \label{fig:attributes}
\end{figure*}

\begin{figure}[tb]
  \centering
  \includegraphics[scale=0.4]{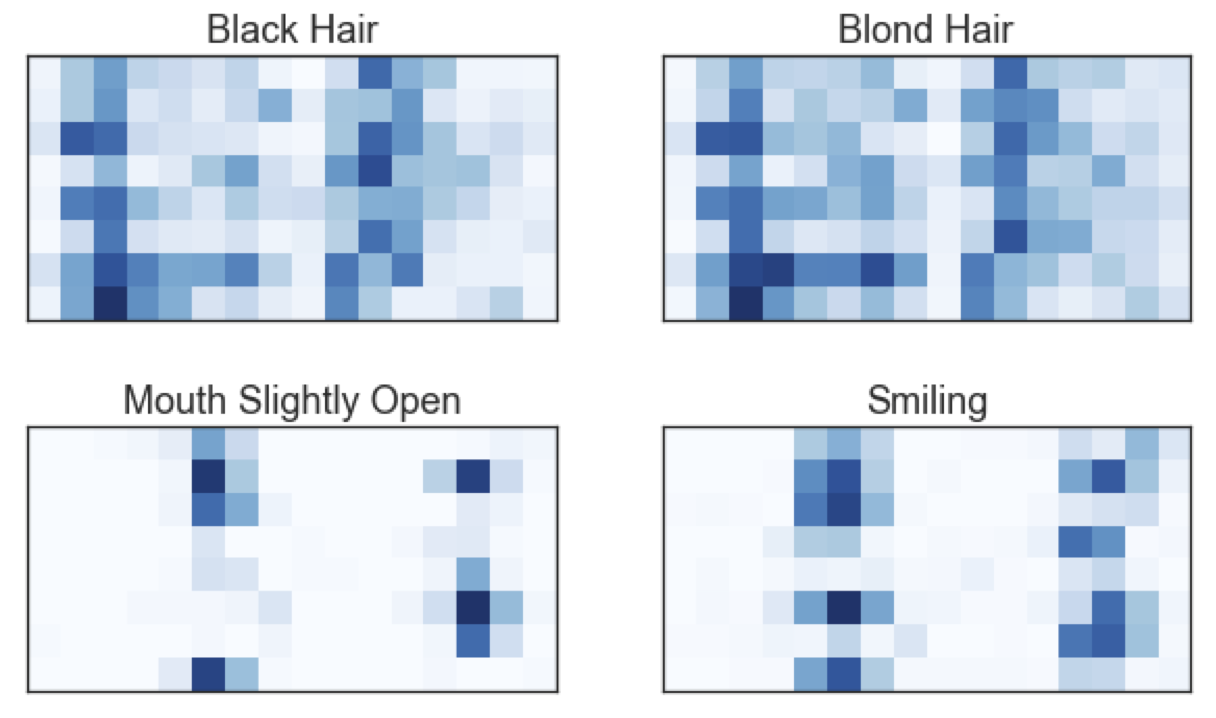}
  \caption{Average of attention mask of each feature. One element represents one-dimensional mean value of attention mask, and there are 128 elements. In this figure, deeper color means higher value.\label{fig:feature_vis}}
\end{figure}

\subsection{Binary Classifiers with Attention Mechanism}
Binary classifiers $F_b = \{f_b^1, f_b^2,\dots,f_b^k \}^K_{k=1}$, where $K$ is number of attributes, are components that perform binary classification corresponding to each attribute of the image.
This network is our main component.
The loss for the binary classifiers can be expressed by a sum of binary cross entropy of each attribute as
\begin{eqnarray}
    L_{b} &=& - \frac{1}{N} \sum^N_{i=1} \sum^K_{k=1} 
    y_{i}^{k} \log{\hat{y}_{b,i}^{k}}
    + (1 - y_{i}^k)\log{(1 - \hat{y}_{b,i}^{k})} \,, 
    \nonumber \\ \\
    \hat{y}_{b,i}^{k} &=& f_b^k((M^k(f_f({\bm x}_i))) \otimes f_f({\bm x}_i))\,,
\end{eqnarray}
where $M^k(\cdot)$ is the attention mask for $k$-th attribute, and $\otimes$ represents element-wise product operation.
Note that the attention mask $M^k$ has the same number of channels as that of feature $f_f$.
It differentiates from existing importance visualization algorithms.

The overall loss function is:
\begin{equation}
    L = L_b + \lambda \cdot \|{\bm M}\|_1\,,
    \label{eq:L}
\end{equation}
where, $\lambda$ is the weight parameter,
and $\|{\bm M}\|_1$ means $L1$ sparseness to the attention mask which is used to extract features that are really important for the data.
Table~\ref{table:binary_classifier} shows the detailed architecture of Binary Classifier that we used in the experiment.

Our main idea is to train sub-networks with multi-channel attention mask for each attribute. 
The attention mask applied to the sub-network is not common in the feature map but has the same number of channels as the feature map. This channel attention mechanism can reveal which channel in the feature map focuses on which part of the image.
This transparency of the feature space leads to predictability for perturbing DNNs.

\renewcommand{\arraystretch}{1.5}
\begin{table}[t]
\centering
\caption{Architecture of Feature Extractor.\label{table:feature_extractor}}
\begin{tabular}{c} \hline\hline
RGB image $x\in\mathbb{R}^{H \times{W}\times{3}}$        \\ \hline
$3\times3$, stride=$1$, padding=$1$, dilation=$1$, conv $32$  \\ \hline
BatchNorm                                      \\ \hline
slope=$0.01$ LeakyReLU                           \\ \hline
$3\times3$, stride=$1$, padding=$2$, dilation=$2$, conv $64$  \\ \hline
BatchNorm                                      \\ \hline
slope=$0.01$ LeakyReLU                           \\ \hline
$3\times3$, stride=$1$, padding=$3$, dilation=$3$, conv $64$  \\ \hline
BatchNorm                                      \\ \hline
slope=$0.01$ LeakyReLU                           \\ \hline
$3\times3$, stride=$1$, padding=$2$, dilation=$2$, conv $128$ \\ \hline
BatchNorm                                      \\ \hline
slope=$0.01$ LeakyReLU                           \\ \hline\hline
\end{tabular}
\end{table}

\begin{table}[t]
\centering
\caption{Architecture of Binary Classifier.\label{table:binary_classifier}}
\begin{tabular}{c} \hline\hline
$z\in\mathbb{R}^{128}$                          \\ \hline
Attention Mask $128$         \\ \hline
Global Average Pooling $128$ \\ \hline
sigmoid                    \\ \hline\hline
\end{tabular}
\end{table}
\renewcommand{\arraystretch}{1}

\subsection{Semantic Channel Pruning}
In this section, we introduce Semantic Channel Pruning as an example of an operation for a human-predictable neural network.
Channel pruning~\cite{he2017channel,ye2018rethinking} is a technique for acquiring sparse DNNs by removing channels in the layers.
Identifying the informative (or important) channels, also known as channel selection, is a key issue
in channel pruning. 
Therefore, we present an example of channel selection using our proposed human-predictable network architecture.

Figure~\ref{fig:feature_semantic_pruning} shows the overview of Semantic Channel Pruning operation.
Our main idea is that we do not need multiple channels for a specific role.
In other words, we assume that if there are multiple channels with similar roles, it is possible to prune all but one.

First, we train the network with the proposed architecture.
Next, we use the trained network and the training data to calculate the correlation matrix among the channels of the network.
Note that the weight of the trained network is fixed at this time.
Finally, we sequentially scan the highly correlated channel pairs and prune the channel with the lower $L1$-norm of the weights.
Our proposed network architecture that accepts this operation satisfies our definition.

\begin{table}[t]
\centering
\caption{Classification accuracy on the CelebA dataset. Our method is not state-of-the-art, but we get enough score for the discussion of explainability.}
\label{table:accuracy_celeba}
\begin{tabular}{l|c} \hline
Method     & Average of accuracy {[}\%{]} \\ \hline\hline
FaceTracer & 81.13                        \\
PANDA-1    & 85.43                        \\
LNet+ANet  & 87.30                        \\
MOON       & 90.93                        \\
ABN        & 91.07                        \\
ResNet101  & 90.69                        \\
Ours       & 90.72                        \\ \hline
\end{tabular}
\end{table}

\subsubsection{Intentional Attention Mask Transformation}
Here, we propose the "Intentional Attention Mask Transformation" to improve the robustness of CNNs.
Our main idea is to reduce the effect of noise by only focusing on important features area for the classification of each attribute.
To achieve this goal, we introduce the following simple attention mask transformation function $g(M; n,\beta)$.
\begin{eqnarray}
    g(M^k; n,\beta) &=& (1 + \beta)\cdot h(M^k;n) - \beta, \\
    h(M^k; n) &=& \begin{cases}
    \frac{(M^k/0.5)^n}{2} & (M^k < 0.5) \\
    1 - \frac{((1-M^k)/0.5)^n}{2} & (M^k \geq 0.5)
    \end{cases}
    \,,
\end{eqnarray}
where $n$ and $\beta$ are parameters to adjust the emphasis and suppression of the mask.
The function $h(M;n)$ is symmetric with respect to 0.5, and the function $g(M; n, \beta)$ emphasizes large values in the mask $M$.

The output of a binary classifier applying our intentional attention mask transformation is:
\begin{equation}
    \hat{y}_{b,i}^{k} = f_b^k(g(M^k(f_f({\bm x}_i); n, \beta)) \otimes f_f({\bm x}_i))\,,
\end{equation}
The above equations emphasize important features about the attribute $k$ of sample $x_i$.

This transformation function is similar to the intensity tone curve in image retouching.
In this paper, the symmetric function inspired by the tone curve is used as the transformation function, and the slope and bias of the symmetric function are used as the side information, but any function and side information (such as prior knowledge of data) can be used.

According to our definition, a model that accepts this operation can be said to be human-predictable because we can know in advance the accuracy change when we change internal parameters.

\begin{table*}[t]
\centering
\caption{Correlation among the attributes. Here, we calculate the correlation among attention masks corresponding to each attribute. It lists the target attributes and the top five attributes that are highly correlated with the target. }
\label{table:correlation_top5}
\begin{tabular}{l|l}
\hline
Target  Attribute & Top5 Highly Correlated Attributes                                                        \\ \hline\hline
Black Hair        & Blond Hair, Brown Hair, Bald, Wearing Hat, Gray Hair                   \\
Heavy Makeup      & Wearing Lipstick, Male, Rosy Cheeks, Attractive, Young                 \\
Bushy Eyebrows    & Bags Under Eyes, Eyeglasses, Arched Eyebrows, Heavy Makeup, Attractive \\
\hline
\end{tabular}
\end{table*}

\section{Experimental Results}
We evaluate our approach using the CelebA dataset \cite{liu2015deep}, which consists of 40 facial attribute labels and 202,599 images (182,637 training images and 19,962 testing images).
The parameter of $L1$ sparseness in Eq.~\ref{eq:L} is $\lambda = 0.00001$. The dimension of the attention mask and feature map is 128.
Tables~\ref{table:feature_extractor} and~\ref{table:binary_classifier} show the details of the network architecture used in this experiment.

% multi-attributed classification results
\begin{comment}
Table~\ref{table:accuracy_celeba} shows the experimental results for the multi-attributed classification.
\end{comment}
Table~\ref{table:accuracy_celeba} shows the overall test accuracies of several networks.
We evaluate the accuracy rate using FaceTracer~\cite{kumar2008facetracer}, PANDA~\cite{zhang2014panda}, LNet+ANet~\cite{liu2015deep}, mixed objective optimization network (MOON)~\cite{rudd2016moon}, Attention Branch Network (ABN)~\cite{fukui2019attention} and ResNet101.
\begin{comment}
Although our network architecture does not outperform the state-of-the-art approach, we can see that it has achieved enough performance in discussing explainability.
\end{comment}
Although the accuracy of our network is not the best, our aim is to increase the interpretability of the network.
In this sense, the performance of the proposed network is reasonably good, even compared with the state-of-the-art network.

\begin{figure}
  \centering
  \includegraphics[scale=0.51]{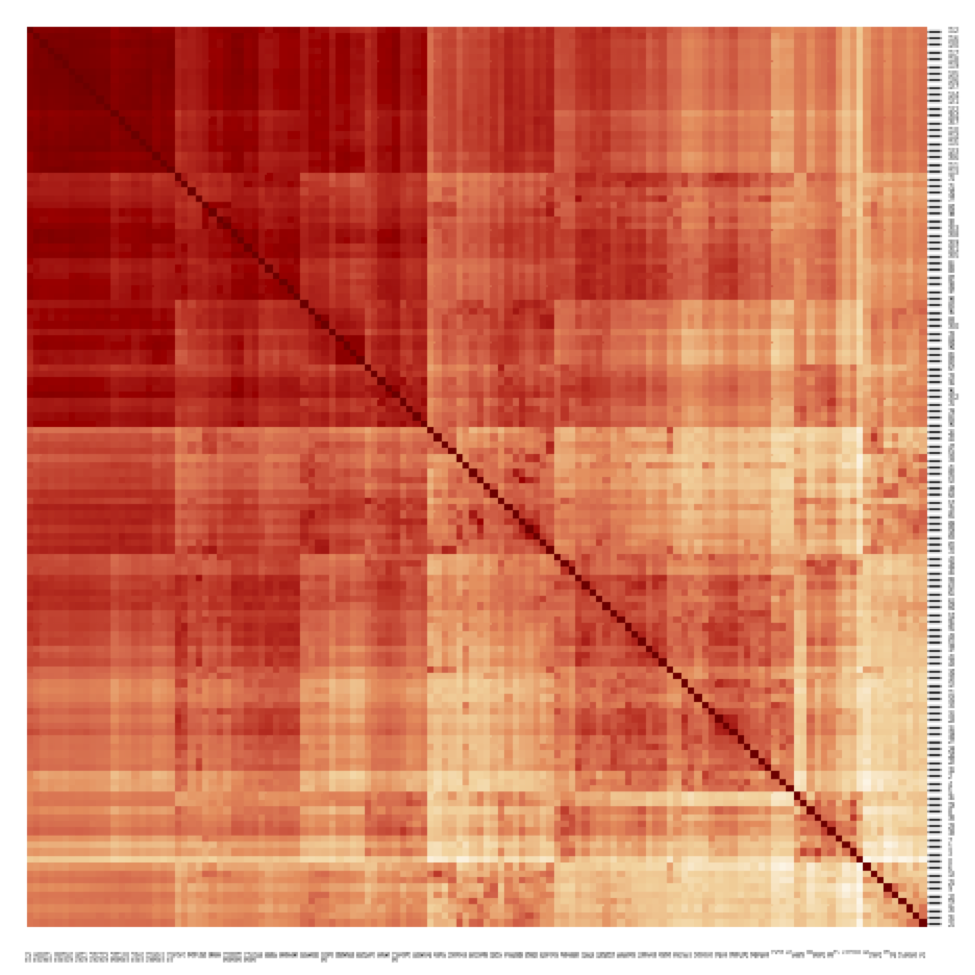}
  \caption{Visualization of correlation in feature space. In this figure, deeper color means higher correlation. In Semantic Channel Pruning operations, highly correlated channels are candidates for pruning.\label{fig:cor_matrix_mask}}
\end{figure}

\begin{figure}[tb]
  \centering
  \includegraphics[scale=0.24]{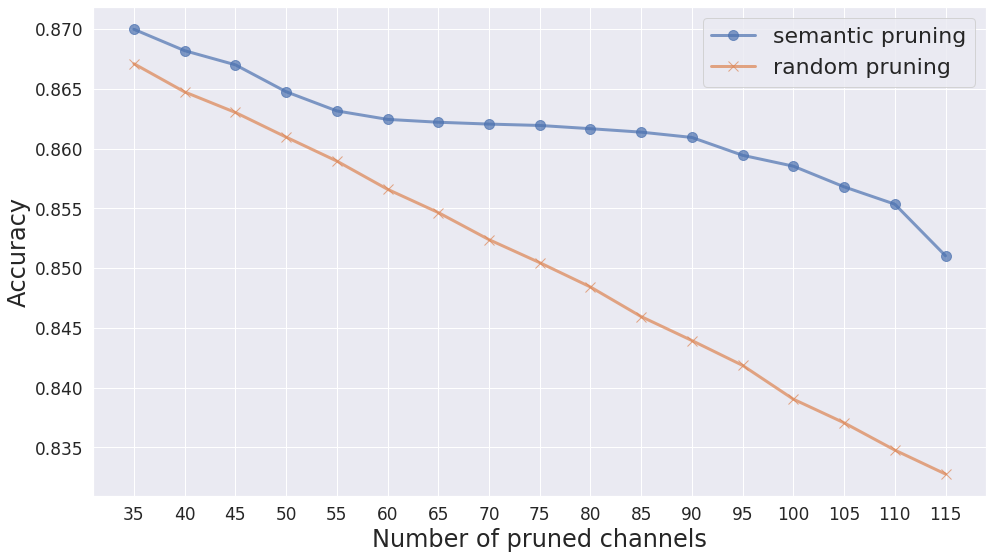}
  \caption{Experimental result of semantic channel pruning operation.
  We gradually increased the number of pruned channels and evaluated the change of accuracy.
  We compare the random selection of the channels with the proposed method and confirm that the proposed method can select more meaningful channels.\label{fig:channel_prune}}
\end{figure}

\begin{comment}
\begin{figure}[tb]
  \centering
  \includegraphics[scale=0.63]{./tsne.png}
  \caption{The result of visualization by TSNE of active and pruned channels after channel pruning.
  We reduced the $64\times 64$-dimensional feature map to $2$ dimensions.
  It can be seen that only the channels that represent each group remained active, and the surrounding channels were subject to pruning.\label{fig:tsne}}
\end{figure}
\end{comment}

Figure~\ref{fig:attributes} shows the visualization of the attention mask by our proposed method. In this figure, common feature channels are visualized for each attribute, and each column means the top three features which have high importance for each attribute. Our attention masks focus on areas that may be important to attributes. In addition, this experimental result suggests that analysis of feature space reveals the relationship among attributes. 
For example, feature IDs 25, 14 and 50 are not used to predict the attribute of Mouse Slightly Open, although they are used for the attribute of Smiling. 
On the other hand, IDs 105 and 50 are commonly used in the attributes of Smiling and Mouse Slightly Open, and ID 8 is not used in Smiling.
Only the mouth region is focused for the attribute of Mouse Slightly Open, while a broad part including mouse and eyes are focused for the attribute of Smiling.
Those results are consistent with human intuition.
In addition, it can be seen that even in different channels, there are cases where attention focus to the same location in the image.

The network architecture we proposed in the previous section can reveal channels that are similar in the role.
Figure~\ref{fig:cor_matrix_mask} shows the visualization of the correlations between features. 

Figure~\ref{fig:feature_vis} shows a grid visualization of the mean values of each channel of attention masks. 
Our analysis reveals that similar channels of features are commonly used for attributes of "Black Hair" and "Blond Hair." Relevant channels for attributes of "Mouth Slightly Open" and "Smiling" are also similar.

Table \ref{table:correlation_top5} lists some of the attributes and their highly correlated attributes.
Here, we calculate the correlation among attention masks corresponding to each attribute.
In other words, considering the data in Figure~\ref{fig:feature_vis} as one vector data, Table~\ref{table:correlation_top5} uses the correlation of the vector data.
Attributes that are intuitively similar are highly correlated. This result makes it possible to group highly correlated attributes. In addition, experimental results may even reveal potential relationships among attributes.
In other words, our network architecture makes it easy for humans to understand changes in the internal structure of the network.

\begin{figure}[t]
  \centering
  \includegraphics[scale=0.23]{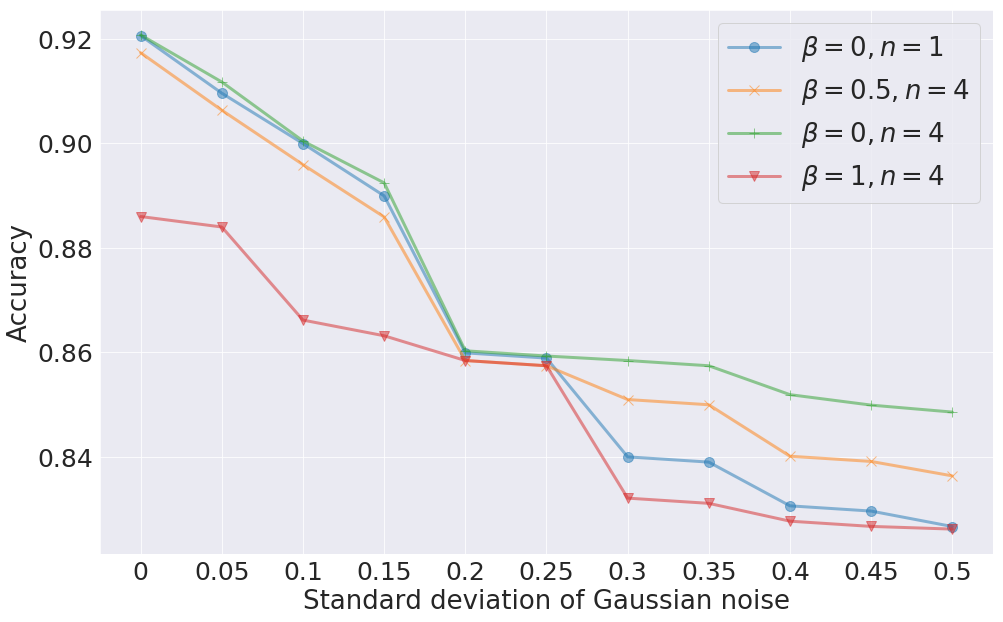}
  \caption{Transition of performance degradation of the network using the intentional attention mask transformation. We add Gaussian noise with a standard deviation of 0 to 0.5 to the input image to create pseudo noisy data. Here, $n$ and $\beta$ are parameters to adjust the emphasis and suppression of the mask.}
  \label{fig:transform}
\end{figure}

\subsection{Channel Pruning Experiment}
Previous results allow us to prune unnecessary channels after understanding the role of each channel in the network.
In our proposed method, the group of channels with the same role becomes the target of pruning.
The channel to be pruned is determined based on the correlation matrix in Figure~\ref{fig:cor_matrix_mask}, that is, the channel pair with high correlation in the correlation matrix is considered as a pair with similar roles, and the channel with the higher $L1$-norm is pruned.

Figure~\ref{fig:channel_prune} shows the experimental result of our Semantic Channel Pruning operation.
We gradually increased the number of channels to be pruned and evaluated the transition of classification performance.
In random pruning, the performance degrades linearly as the number of pruned channels increases.
On the other hand, we can see that our proposed channel pruning method selects meaningful channels compared to random selection.

\begin{comment}
In addition, Figure~\ref{fig:tsne} shows the results of visualizing the active and pruned channels using TSNE~\cite{maaten2008visualizing} after channel pruning.
We reduced the $64\times 64$-dimensional feature map to $2$ dimensions.
In this visualization, $100$ out of $128$ channels were pruned.
It can be seen that only the channels that represent each group remained active, and the surrounding channels are subject to pruning.
\end{comment}

\subsection{Mask Transformation Experiment}
We evaluate the robustness of the network by the Intentional Attention Mask Transformation.
We add Gaussian noise with a standard deviation of 0 to 0.5 to the input images to create pseudo noisy data.
Figure~\ref{fig:transform} shows the experimental results of observing the transition of the performance for this noisy input data for a combination of multiple parameters.
The experimental results show that the transformation of the attention mask makes the network robust against performance degradation due to noise. The result of $\beta = 0$ and $n = 1$ is the performance transition of the network without the intentional attention mask transformation, and by adjusting $\beta$ and $n$, the performance improvement for noisy input is achieved.

\section{Conclusion}
In this paper, We proposed the novel aspect of the interpretability of DNNs.
We define "human predictability of the model" as "easiness for the human to predict the change of the inference when perturbing the model."
This means that even if the function is complex, the result of the operation on it must match human intuition.
Furthermore, we propose an example of a neural network architecture and operation pair that satisfies our definition.

We suggest that considering various applications according to our definition will lead to a wide range of research.
As one of the future works, we need to find more model-operation pairs that satisfy our definition and show the potential for the broader application of our definition.

% \section{Acknowledgments}

\bibliographystyle{aaai}

\end{document}